\documentclass[10pt,twocolumn,letterpaper]{article}

\usepackage[pagenumbers]{cvpr} 

\usepackage{graphicx}
\usepackage{amsmath}
\usepackage{amssymb}
\usepackage{booktabs}
\usepackage{enumerate}

\usepackage{amsfonts}
\usepackage{amsthm}
\usepackage{color}
\usepackage{algorithm,algpseudocode}
\usepackage{multicol}
\usepackage{array,multirow}
\newcommand{\tr}{\mathrm{Tr}}

\newtheorem{proposition}{Proposition}
\usepackage{multirow}
\usepackage{tabularx}
\usepackage{listings}
\usepackage{xcolor}
\usepackage{color}
\usepackage{colortbl}
\definecolor{highlight}{gray}{0.9}
\definecolor{kqcolor}{RGB}{219,90,107}

\makeatletter
\DeclareRobustCommand\onedot{\futurelet\@let@token\@onedot}
\def\@onedot{\ifx\@let@token.\else.\null\fi\xspace}

\def\eg{\emph{e.g}\onedot} 
\def\ie{\emph{i.e}\onedot}

\makeatother

\usepackage{hyperref}
\hypersetup{pagebackref=true,breaklinks=true,colorlinks,bookmarks=false}

\usepackage[capitalize]{cleveref}
\crefname{section}{Sec.}{Secs.}
\Crefname{section}{Section}{Sections}
\Crefname{table}{Table}{Tables}
\crefname{table}{Tab.}{Tabs.}

\begin{document}

\title{Mimicking the Oracle: An Initial Phase Decorrelation Approach for Class Incremental Learning}
\author{Yujun Shi\textsuperscript{\rm 1}
\quad\quad
Kuangqi Zhou\textsuperscript{\rm 1}
\quad\quad
Jian Liang\textsuperscript{\rm 3}
\quad\quad
Zihang Jiang\textsuperscript{\rm 1} \\
Jiashi Feng\textsuperscript{\rm 2}
\quad\quad
Philip Torr\textsuperscript{\rm 4}
\quad\quad
Song Bai\textsuperscript{\rm 2}
\quad\quad
Vincent Y. F. Tan\textsuperscript{\rm 1} \\
\textsuperscript{\rm 1}National University of Singapore
\quad\quad \textsuperscript{\rm 2} ByteDance Inc.\\
\textsuperscript{\rm 3} Institute of Automation, Chinese Academy of Sciences (CAS)
\quad\quad
\textsuperscript{\rm 4} University of Oxford \\
{\tt\small shi.yujun@u.nus.edu} \quad {\tt\small vtan@nus.edu.sg}
}
\maketitle

\begin{abstract}
Class Incremental Learning (CIL) aims at learning a classifier in a phase-by-phase manner, in which only data of a subset of the classes are provided at each phase.
Previous works mainly focus on mitigating forgetting in phases after the initial one.
However, we find that improving CIL at its {\em initial phase} is also a promising direction.
Specifically, we experimentally show that directly encouraging CIL Learner at the initial phase to output similar representations as the model jointly trained on all classes can greatly boost the CIL performance.
Motivated by this, we study the difference between a na\"ively-trained initial-phase model and the oracle model.
Specifically, since one major difference between these two models is the number of training classes, we investigate how such difference affects the model representations.
We find that, with fewer training classes, the data representations of each class lie in a long and narrow region; with more training classes, the representations of each class scatter more uniformly.
Inspired by this observation, we propose \emph{Class-wise Decorrelation (CwD)} that effectively regularizes representations of each class to scatter more uniformly, thus mimicking the model jointly trained with all classes (\ie, the oracle model).
Our CwD is simple to implement and easy to plug into existing methods.
Extensive experiments on various benchmark datasets show that CwD consistently and significantly improves the performance of existing state-of-the-art methods by around 1\% to 3\%.
Code: \href{https://github.com/Yujun-Shi/CwD}{https://github.com/Yujun-Shi/CwD}.
\end{abstract}

\begin{figure}[htbp]
\centering
\includegraphics[width=8cm]{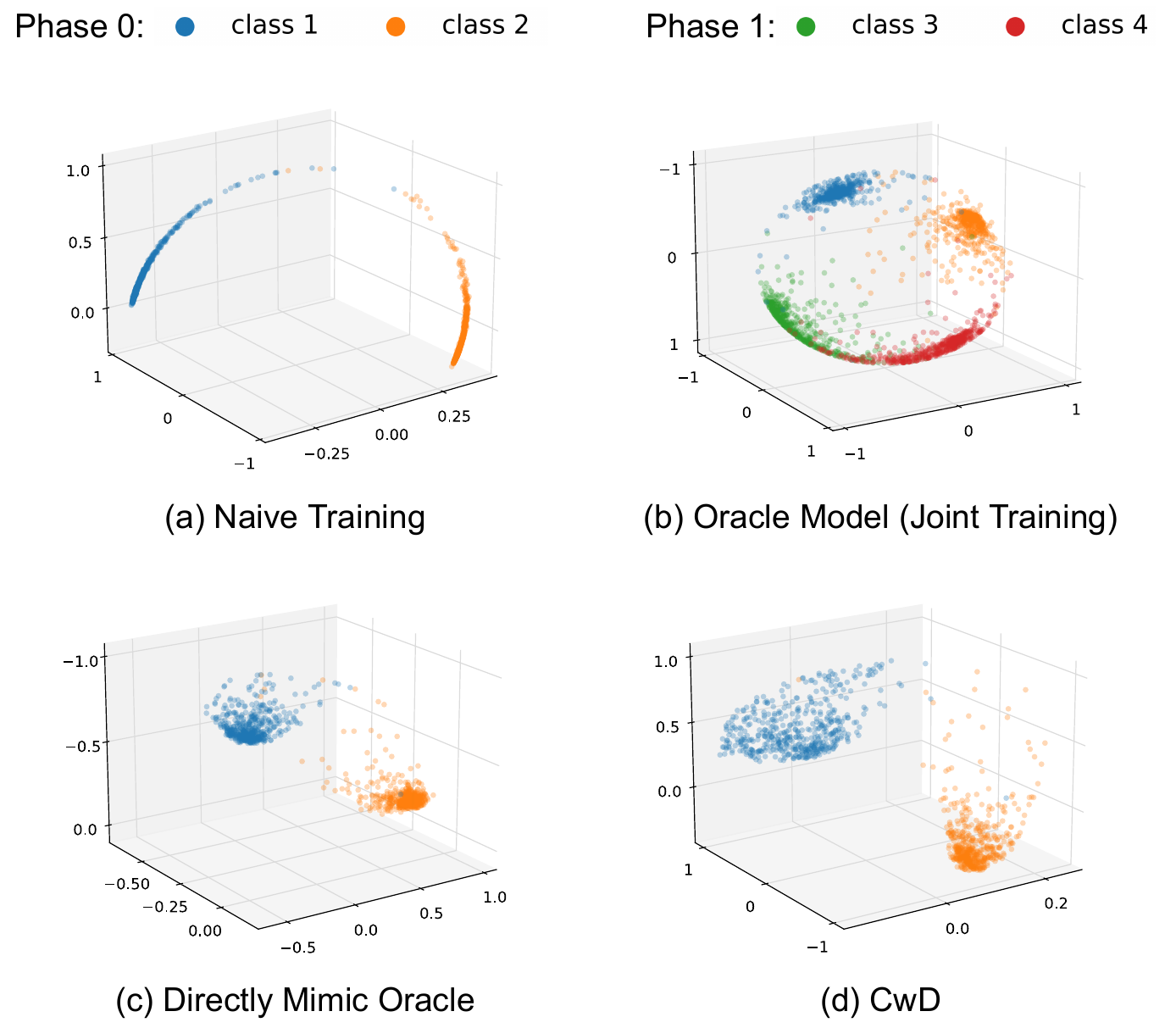}
\vspace{-2mm}
\caption{Visualization of representations (normalized to the unit sphere) in a two-phase CIL setting (learning 2 classes for each phase).
\textbf{(a)} Na\"ive training at the initial phase (a.k.a., phase~$0$). The data representations of each class lie in a long and narrow region.
\textbf{(b)} Joint training on all 4 classes (oracle model). The data representations of each class scatter more uniformly.
\textbf{(c)} Directly mimicking the oracle model at the initial phase, \ie, training the CIL learner with a regularization term that enforces the learner to output representation that is similar to the oracle model. This makes the representations of each class scatter more uniformly (like (b)).
\textbf{(d)} Training at the initial phase with our CwD regularizer, which also yields uniformly-scattered representations (like (b) and~(c)).
Best viewed in color.
}
\label{figure1}
\vspace{-3mm}
\end{figure}

\section{Introduction}
The ability to continually acquire new knowledge is a key to achieve artificial intelligence.
To enable this ability for classification models, \cite{rebuffi2017icarl,hou2019learning,wu2019large,douillard2020podnet,liu2020mnemonics,liu2021adaptive,simon2021learning} introduce and study {\em Class Incremental Learning} (CIL).
In CIL, training is conducted phase-by-phase, and only data of a subset of classes are provided at each phase.
The goal of CIL is to perform well on classes learned at the current phase as well as all previous phases.

The major challenge of CIL is that the model performance on previously learned classes usually degrades seriously after learning new classes, a.k.a. catastrophic forgetting~\cite{mccloskey1989catastrophic,french1999catastrophic}.
To reduce forgetting, most previous works~\cite{li2017learning,rebuffi2017icarl,hou2019learning,douillard2020podnet,simon2021learning} focus on phases after the initial one, \eg introducing forgetting-reduction regularization terms that enforce the current-phase model and the previous-phase model to produce similar outputs of the same input.

However, the role of the initial phase in CIL (the phase before the CIL learner begins incrementally learning new classes) is largely neglected and much less understood.
We argue that the initial phase is of critical importance, since the model trained at this phase implicitly affects model learning in subsequent CIL phases (\eg, through the forgetting-reduction regularization term).
In this work, we thus study whether and how we can boost CIL performance by improving the representations of the initial phase.

To start with and to motivate our method, we conduct an exploratory experiment to investigate the potential of improving CIL at its initial phase.
Specifically, at the initial phase, we regularize the CIL learner to produce similar representations as the model trained with data of all classes (\ie, the oracle model), since the upper bound of CIL is the oracle model.
According to our results, this additional regularization drastically improves CIL performance.
In addition, as we experimentally show that, although this term is used in the initial phase, it yields little performance gain in the initial phase. In contrast, it significantly benefits CIL performance in subsequent phases.
This demonstrates that the performance improvements are not simply due to a higher accuracy at the initial phase, but because this regularization makes the initial-phase representations more favorable for incrementally learning new classes.

Inspired by this, we consider improving CIL from a novel perspective---encouraging the CIL learner to mimic the oracle model in the initial phase.
To achieve this, we first need to understand the difference between representations produced by a na\"ively-trained initial-phase model and the oracle model.
Specifically, since the oracle model is trained with more classes, we investigate how representations are affected by the number of training classes.
To this end, we compute and analyze the eigenvalues of the covariance matrix of representations of each class.
Interestingly, we find that when training with fewer classes, the top eigenvalues of the covariance matrix of representations of each class dominate, indicating that the representations of each class lie in a long and narrow region (see Fig.~\ref{figure1}~(a) for example).
On the other hand, for models trained with more classes (particularly, the oracle model), the top eigenvalues become less dominant, indicating that the representations of each class scatter more uniformly (see Fig.~\ref{figure1}~(b)).

We are thus motivated to enforce data representations of each class to be more uniformly scattered at the initial phase, which mimics the representations produced by the oracle model.
To this end, we first theoretically show that, a group of embeddings will scatter more uniformly in the space if its correlation matrix has smaller Frobenius norm. We then propose to minimize the Frobenius norm of the correlation matrix of the data representations for each class.
We refer to our regularization term as \emph{Class-wise Decorrelation (CwD)}.
We provide a visualization to summarize our motivation and methodology in Fig.~\ref{figure1}.
Our proposed CwD regularizer can serve as a generic plug-in to other CIL methods and can be easily implemented.

Extensive experiments on various benchmark datasets show that our CwD regularizer works well with state-of-the-art CIL methods, yielding significant and consistent performance gain in different settings.
In addition, we also perform detailed ablation studies on how the effectiveness of CwD is influenced by factors such as the number of classes at the initial CIL phase, the number of exemplars for each class and regularization coefficient of the CwD term.

The contributions of this paper are as follows: 1) We empirically discover that encouraging the CIL learner to mimic the oracle model in the initial phase can boost the CIL performance. 2) We find that compared with na\"ively-trained initial-phase model, data representations of each class produced by the oracle model scatter more uniformly, and that mimicking such representations at the initial phase can benefit CIL. 3) Based on our findings, we propose a novel \emph{Class-wise Decorrelation (CwD)} regularization technique to enforce representations of each class to scatter more uniformly at the initial CIL phase. 4) Extensive experiments show that our proposed CwD regularization yields consistent improvements over previous state-of-the-art methods.

\section{Related Works}
Two classic setups of Incremental Learning are Class Incremental Learning (CIL)\cite{rebuffi2017icarl,hou2019learning,liu2020mnemonics,liu2021adaptive,hu2021distilling,ahn2020ss,masana2020class} and Task Incremental Learning (TIL)\cite{kirkpatrick2017overcoming,lopez2017gradient,aljundi2018memory, shi2021continual,wang2021training,tang2021layerwise,mirzadeh2020understanding}. CIL and TIL both split all training classes into multiple tasks and learn them sequentially. 
The difference between these two setups is that TIL allows using task information during inference (\ie, knowing what task does test data belong to) but the CIL does not. In this work, we focus on the setting of CIL.
The major challenge of CIL is that model performance on previously learned classes degrades drastically after learning new classes, \ie catastrophic forgetting \cite{mccloskey1989catastrophic,french1999catastrophic}.

Many CIL methods mitigate forgetting through knowledge distillation\cite{li2017learning,rebuffi2017icarl,hou2019learning,douillard2020podnet,simon2021learning}. In these methods, when learning at a new phase, the model of the previous phase is used as the teacher, and the CIL Learner is regularized to produce similar outputs as  the teacher. In this way, the knowledge of previously learned classes can be preserved.

However, distillation-based methods introduce the dilemma of balancing between previously learned classes and current classes. In particular, if the distillation term is too large, then the model's ability to learn new classes will be limited. In contrast, if the distillation term is too small, forgetting will be amplified. To mitigate this dilemma, some methods have been proposed to maintain a good balance between old and new classes  \cite{wu2019large,wu2021striking,belouadah2019il2m,liu2021adaptive}.

The common focus of existing methods is on improving CIL at phases after the initial one. Differently, we study CIL from a less-explored perspective --- improving CIL at its initial-phase representations. 
Previously, \cite{Mittal_2021_CVPR} also studied the initial-phase representation for CIL.
However, their work focuses on the relation between over-fitting at the initial phase and CIL performance. Their main observation is that leveraging some known generalization-improving techniques (\eg, heavy-aug\cite{cubuk2019autoaugment}, self-distillation\cite{furlanello2018born}) can improve CIL.
By contrast, in our work, based on the novel observation that mimicking the oracle model at the initial phase is beneficial, we reveal how number of training classes affects model representations, and further propose our Class-wise Decorrelation (CwD).
Unlike techniques such as self-distillation\cite{furlanello2018born} and heavy-aug\cite{cubuk2019autoaugment} that brings significant higher accuracy at initial phase, our CwD benefits CIL mainly by making the initial-phase representation more favorable for incrementally learning new classes.

Feature decorrelation has also been explored in some other research fields. For example, \cite{zbontar2021barlow,hua2021feature,bardes2021vicreg} rely on feature decorrelation to solve the mode collapse problem in self-supervised learning when negative pairs are missing, and \cite{huang2018decorrelated,cogswell2015reducing,xiong2016regularizing} use decorrelation to improve neural networks' features and hence boost generalization.
Differently, our work focuses on CIL, and propose to use class-wise feature decorrelation to mimic the oracle model.

\cite{roth2020revisiting} uses eigenvalue analysis on representations of Deep Metric Learning (DML) models. They find that preventing representations to be overly compressed can improve DML generalization. They achieve this by randomly switching negative samples with positive samples in the ranking loss. Differently, our work focuses on CIL, and propose to use class-wise feature decorrelation, which is a more effective way to counter compression of representations.

\section{Methodology}
In this section, we investigate boosting CIL performance by improving the initial-phase representations. This strategy is different from most previous works. 

Firstly, in Sec.~\ref{oracle_init}, we investigate the potential of improving CIL through mimicking the oracle model representations at the initial phase.

Motivated by the observation made, in Sec.~\ref{visualization}, we conduct an eigenvalue analysis on covariance matrix of representations of each class to study how the number of classses used for training affects representations.

We then further develop a novel regularization term, namely the \emph{Class-wise Decorrelation (CwD)} in Sec.~\ref{class_wise_decorr}. We show mathematically and experimentally that this regularization term is effective in enforcing data representations of each class to scatter more uniformly.

\subsection{Directly Mimicking the Oracle Model Representation at Initial Phase Improves CIL}
\label{oracle_init}
In this section, we conduct an exploratory experiment to see whether encouraging the CIL learner to directly mimic the oracle model at the initial phase can improve performance.

Specifically, at the initial CIL phase, we add an additional regularization term to encourage the model to output similar representations as a oracle model, yielding the following objective function:
\begin{equation}
\label{oracle_obj}
   \min_{\theta} L_{\mathrm{ce}}(x,y,\theta) + \beta \left(1 - \frac{f_{\theta}(x)^{\top}f_{\theta^{*}}(x)}{\|f_{\theta}(x)\|_{2}\|f_{\theta^{*}}(x)\|_{2}}\right),
\end{equation}
where $\theta$ denotes the model parameters, and $\theta^{*}$ denotes parameters of the oracle model (which are fixed).
$L_{\mathrm{ce}}(x,y,\theta)$ is the standard cross entropy loss, $(x,y)$ is the input training data-label pair, and $\beta$ is a hyper-parameter controlling the strength of the regularization. $f_{\theta}(x)$ and $f_{\theta^{*}}(x)$ denote the representations produced by the CIL learner and the oracle model, respectively.
The second term in this objective is the regularization that enforces $f_{\theta}(x)$ to be similar to $f_{\theta^{*}}(x)$.

\begin{figure}[htbp]
\centering
    \includegraphics[width=8.3cm]{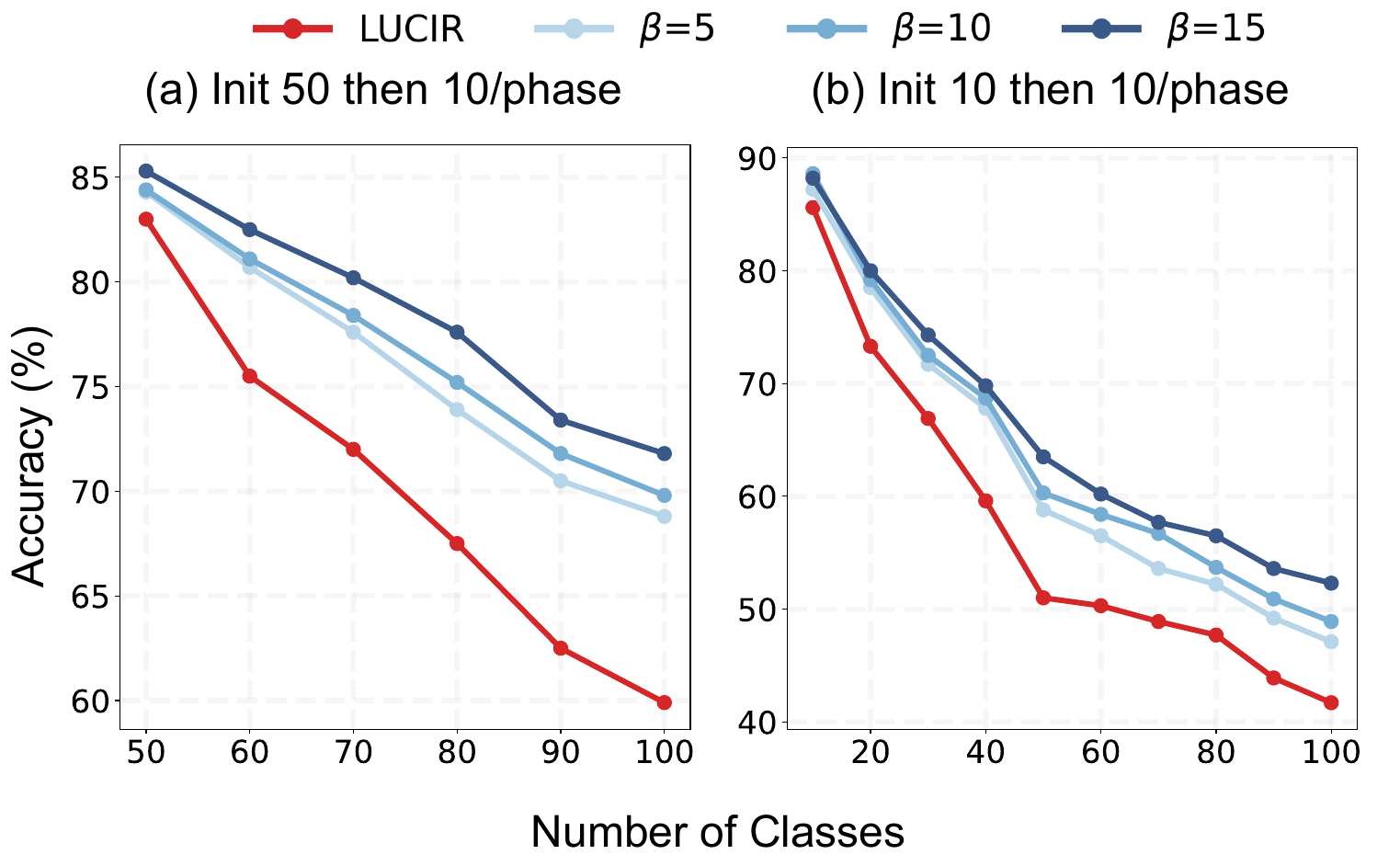}
    \vspace{-6mm}
    \caption{\textbf{The effectiveness of directly the mimicking the representations of the oracle model at the initial phase}. (a) Initially trained on 50 classes, and then incremented with  10 classes per phase for 5 more phases. (b) Initially trained on 10 classes and then incremented with 10 classes per phase for 9 more phases. The regularization coefficient $\beta$ is defined in  Eqn.~\eqref{oracle_obj}. We show the accuracy of each CIL phases. Results are averaged over 3 runs.}
    \label{oracle_results}
    \vspace{-3mm}
\end{figure}

We experiment on the following two protocols with ImageNet100 and ResNet18 \cite{he2016deep}: (1) the CIL learner is initially trained on 50 classes and then incremented with 10 classes per phase for 5 more phases; (2) the CIL learner is initially trained on 10 classes and then incremented 10 classes per phase for 9 more phases.
Under these two protocols, we use Eqn.~\eqref{oracle_obj} as the optimization objective for the initial phase with the strong baseline of LUCIR~\cite{hou2019learning}.
For the following phases, no regularization is added and the original LUCIR\cite{hou2019learning} is applied.
As can be observed in Fig.~\ref{oracle_results}, the regularization term at the initial phase can greatly improves CIL performance.
Notably, in the second protocol, although only 10 classes are used at the initial phase, this regularization still brings significant improvements.
In addition, in both protocols, although this regularization is only applied at the initial phase, it negligibly improves the accuracy of initial phase, but significantly improves the performance in subsequent phases.
This would demonstrate that the improvements are not due simply to an accuracy boost at the initial phase, but because the initial-phase model is more favorable for incrementally learning new classes.

Since the oracle model is not directly available in practice, in the following sections, we explore the characteristics of the oracle model representations and try improving CIL by mimicking this characteristic.

\subsection{Class-wise Representations of Oracle Model Scatter More Uniformly}
\label{visualization}
Motivated by the significant improvements yielded by mimicking the oracle model at the initial phase, we investigate the difference between na\"ively-trained initial-phase model and the oracle.

Specifically, since the oracle model is trained with more classes than na\"ively-trained initial-phase model, we conduct an eigenvalue analysis to understand how the number of classes used for training affects representations.
Using ImageNet100, we generate four subsets containing 10/25/50/100 classes, where the subset with more classes contains the subset with fewer classes (the 10 classes of the first subset are shared by all 4 subsets).
We train four ResNet18 models on each of the subset, and analyze the difference on the representations.

The details of our eigenvalue analysis are elaborated as follows.
For a given class $c$, suppose we have $n$ data points, we denote $Z^{(c)}_{i} \in \mathbb{R}^{d}$ as the model output representation on the $i$-th data point of class $c$,
and the mean vector of all representations of class $c$ is denoted as $\bar{Z}^{(c)} = \frac{1}{n}\sum_{i=1}^n Z_i^{(c)}$.
The covariance matrix of class $c$'s representations is estimated in an unbiased manner as   
\begin{equation}
\label{cov_mat}
   K^{(c)} = \frac{1}{n-1} \sum_{i=1}^{n} (Z^{(c)}_{i} - \bar{Z}^{(c)})(Z^{(c)}_{i} - \bar{Z}^{(c)})^{\top}.
\end{equation}
Based on the estimated covariance matrix of class $c$'s representations, we perform an eigendecomposition  $K^{(c)} = U \Sigma^{(c)} U^{\top}$, where $\Sigma^{(c)}$ is a diagonal matrix with eigenvalues $(\lambda^{(c)}_{1}, \lambda^{(c)}_{2},\ldots,\lambda^{(c)}_{d})$ on the diagonal. Without loss of generality, we assume that the eigenvalues are sorted in descending order.
To observe whether the top eigenvalues dominate, we define
\begin{equation}
\label{alpha_k}
   \alpha^{(c)}_{k} := \frac{\sum_{i=1}^{k}\lambda^{(c)}_{i}}{\sum_{i=1}^{d}\lambda^{(c)}_{i}} \in [0,1],
\end{equation}
which measures the proportion of variance represented by the top $k$ eigenvalues.
If $\alpha_{k}^{(c)}$ is close to $1$ even when $k$ is small, then the top eigenvalues of $K^{(c)}$ dominate.

For one of the 10 shared classes among the four models, we visualize how $\alpha^{(c)}_{k}$ changes with increasing $k$.
Results on representations of other classes show similar trend, and are in the Appendix.
As can be observed in Fig.~\ref{svd_analysis}, for the model trained with only 10 classes, $\alpha^{(c)}_{k}$ increases quickly for $k \in \{1,2,\ldots, 10\}$, and then saturates at a value close to 1 as $k$ keep increasing.
This shows that for the 10 class model, the top eigenvalues dominate for covariance matrix of data representations of each class, indicating that data representations lie in a long and narrow region.
In addition, for any fixed $k$, $\alpha^{(c)}_{k}$ strictly decreases as the model is being trained with more classes.
This shows that, as the model is trained with more classes, the top $k$ eigenvalues become less dominant, suggesting that the data representations of each class scatter more uniformly.

Since the oracle model is trained with more classes than the na\"ively-trained initial-phase model, class-wise data representations of the oracle model scatter more uniformly.

\begin{figure}
   \centering
   \includegraphics[width=7cm]{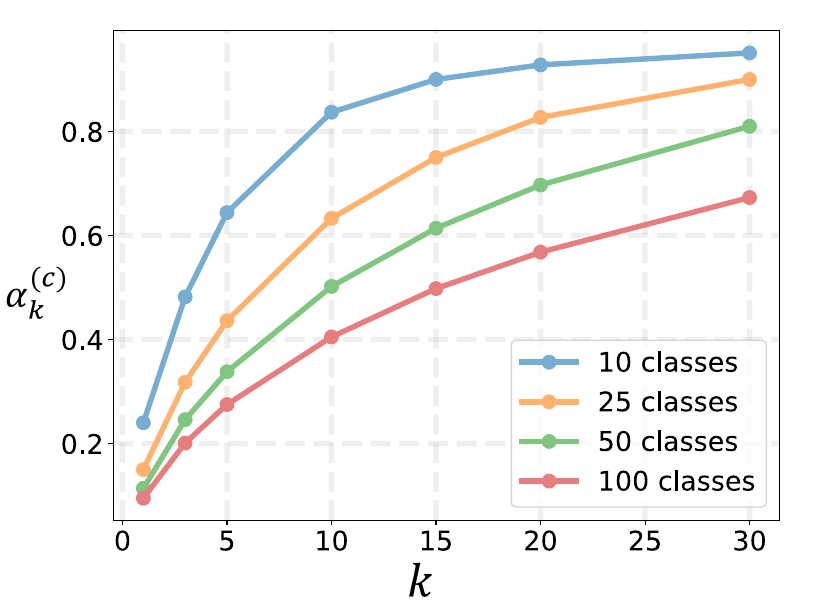}
   \vspace{-4mm}
   \caption{\textbf{Visualization on how $\alpha^{(c)}_{k}$ changes with increasing $k$ for models trained with different number of classes.} $\alpha^{(c)}_{k}$, which measures the proportion of variance represented by the top $k$ eigenvalues, is defined in Eqn.~\eqref{alpha_k}. We plot curve of $\alpha^{(c)}_{k}$ for ResNet18 models trained with 10/25/50/100 ImageNet classes.
   }
   \label{svd_analysis}
   \vspace{-3mm}
\end{figure}

\begin{figure*}
   \centering
   \includegraphics[width=16cm]{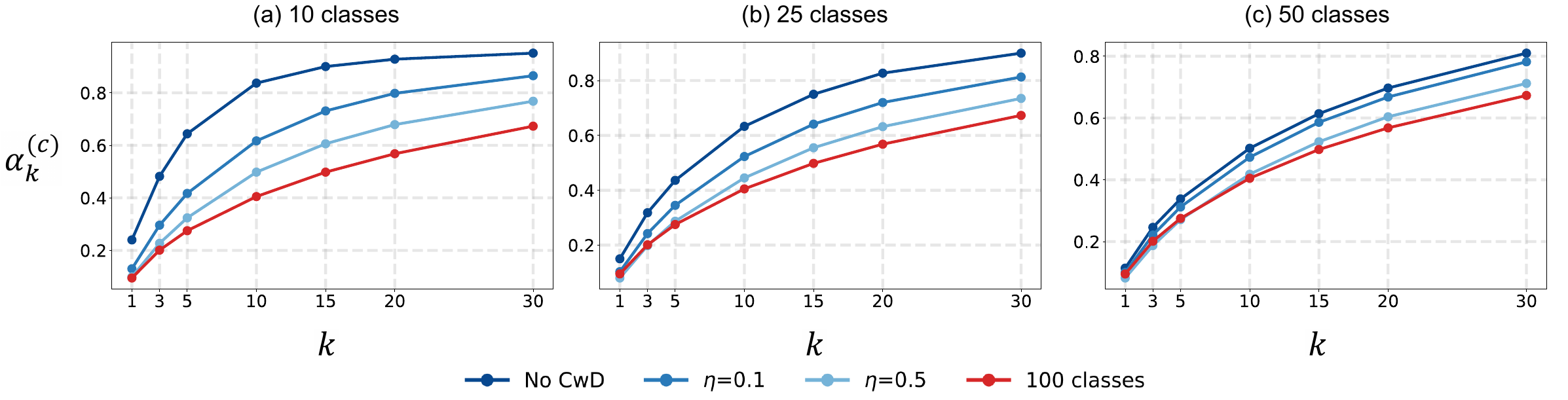}
   \vspace{-3mm}
   \caption{\textbf{Effects of class-wise decorrelation on representations of each class.} $\alpha^{(c)}_{k}$, which measures the proportion of variance represented by the top $k$ eigenvalues, is defined in Eqn.~\eqref{alpha_k}. $\eta$ is the CwD coefficient used in Eqn.~\eqref{overall_obj}. We plot curve of $\alpha^{(c)}_{k}$ with or without CwD objective when training with 10/25/50 classes. We also plot the curve for the model trained on all 100 classes for comparison.}
   \label{decorr_svd_analysis}
   \vspace{-3mm}
\end{figure*}

\subsection{Class-wise Decorrelation}
\label{class_wise_decorr}
The observation in Sec.~\ref{visualization} suggests that one way to encourage the CIL learner to mimic the oracle model at the initial phase is to enforce data representations of each class to scatter more uniformly.

This can be achieved by adding the following regularization objective for each class~$c$ in the initial phase:
\begin{equation}
   \label{l_shape_def}
   L^{(c)}_{\mathrm{shape}} = \frac{1}{d}\sum_{i=1}^{d}\bigg(\lambda^{(c)}_{i} - \frac{1}{d}\sum_{j=1}^{d}\lambda^{(c)}_{j}\bigg)^{2},
\end{equation}
where $d$ is dimension of the representation space.
Minimizing this objective will enforce all eigenvalues to be close, preventing the top eigenvalues to dominate and encouraging the data representations of class $c$ to scatter more uniformly. However, this regularization objective is not practical because computing eigenvalues is expensive.

In order to achieve our desired regularization in an implementation-friendly way, we first normalize all representations of class $c$ by
\begin{equation}
   Z_{i}^{(c)} := \frac{Z_{i}^{(c)} - \bar{Z}^{(c)}}{\sigma^{(c)}(Z)},
\end{equation}
where $\sigma^{(c)}(Z) \in \mathbb{R}^{d}$ is the vector of standard deviations all the representations, and the division is done element-wise.
This normalization results in the covariance matrix $K^{(c)}$ (defined in Eqn.~\eqref{cov_mat}) being equivalent to a {\em correlation matrix}, which satisfies
\begin{equation}
    \sum_{j=1}^{d}\lambda^{(c)}_{j} = \tr(K^{(c)}) = d,
\end{equation}
where $\tr(\cdot)$ is the matrix trace operator and $d$ is the dimension of $K^{(c)}$.

Then, by the following proposition, we can relate the Frobenius norm of a correlation matrix and its eigenvalues.
\begin{proposition}
For a $d$-by-$d$ correlation matrix $K$ and its eigenvalues $(\lambda_{1},\lambda_{2}, \ldots ,\lambda_{d})$, we have:
\begin{equation}
   \sum_{i=1}^{d}\bigg(\lambda_{i} - \frac{1}{d}\sum_{j=1}^{d}\lambda_{j}\bigg)^{2} = \|K\|_{\mathrm{F}}^{2} - d.
\end{equation}
\end{proposition}
The proof of this proposition is given in the Appendix.
It shows that for any correlation matrix $K$, minimizing $L_{\mathrm{shape}}$ defined in Eqn.~\eqref{l_shape_def} is equivalent to minimizing $\|K\|^{2}_{\mathrm{F}}$.

With this proposition, we convert the impractical regularization in Eqn.~\eqref{l_shape_def} into the \emph{Class-wise Decorrelation (CwD)} objective below, which penalizes $\|K^{(c)}\|^{2}_{\mathrm{F}}$ for every class $c$:
\begin{equation}
   L_{\mathrm{CwD}}(\theta) = \frac{1}{C\cdot d^{2}}\sum_{c=1}^{C}\|K^{(c)}\|^{2}_{\mathrm{F}},
\end{equation}
where $C$ is the number of classes used when training at the initial phase, $K^{(c)}$ is the correlation matrix of class $c$ estimated over the training data batch. Note that $K^{(c)}$ is a function of the parameters $\theta$ through its eigenvalues $\lambda_i$.

Therefore, the overall optimization objective at the initial phase is:
\begin{equation}
\label{overall_obj}
    \min_{\theta} L_{\mathrm{ce}}(x,y,\theta) + \eta\cdot L_{\mathrm{CwD}}(\theta),
\end{equation}
where $\eta$ is the hyper-parameter controlling strength of our CwD objective.
A Pytorch-style pseudocode for our proposed CwD regularization is given in Algorithm~\ref{pseudocode}.

\begin{algorithm}[tbp]
   \caption{PyTorch-style pseudocode for CwD.}
   \label{pseudocode}
   
    \definecolor{codeblue}{rgb}{0.25,0.5,0.5}
    \lstset{
      basicstyle=\fontsize{7.2pt}{7.2pt}\ttfamily\bfseries,
      commentstyle=\fontsize{7.2pt}{7.2pt}\color{codeblue},
      keywordstyle=\fontsize{7.2pt}{7.2pt},
    }
\begin{lstlisting}[language=python]
# N: batch size
# d: representation dimension
# z: a batch of representation, with shape (N, d)
# y: a batch of label corresponding to z
def class_wise_decorrelation_loss(z, y):
    loss_cwd = 0.0 # initialize cwd loss
    unique_y = y.unique() # all classes in the batch
    for c in unique_y:
        # obtain all representation of class c
        z_c = z[y==c, :]
        N_c = z_c.size(0)

        # skip if class c only have 1 sample
        if N_c == 1:
            continue

        # normalize representation as in eq.(5)
        z_c = (z_c - z_c.mean(0)) / z_c.std(0))
        # estimate correlation matrix
        corr_mat = 1/(N_c-1)*torch.matmul(z_c.t(), z_c)
        # calculate CwD loss for class c
        loss_cwd += (corr_mat.pow(2)).mean()
    return loss_cwd
\end{lstlisting}
\end{algorithm}

To empirically verify that our proposed $L_{\mathrm{CwD}}$ is indeed effective in encouraging representations of each class to scatter more uniformly, we conduct the same eigenvalue analysis as in Sec.~\ref{visualization}. 
We apply $L_{\mathrm{CwD}}$ to conduct experiments with the same $10/25/50$-class settings in Sec.~\ref{visualization}.
As can be seen in Fig.~\ref{decorr_svd_analysis}, applying $L_{\mathrm{CwD}}$ can effectively decrease $\alpha^{(c)}_{k}$ for every fixed $k$ and every model, and using larger $\eta$ will decrease $\alpha^{(c)}_{k}$ more.
These observations show that the data representations of each class scatter more uniformly after applying $L_{\mathrm{CwD}}$.

\definecolor{hl}{gray}{0.9}
\renewcommand{\arraystretch}{1.1}{
\setlength{\tabcolsep}{1.0mm}{
\begin{table*}[ht]
\centering
\begin{tabular}{lcccccccccc}
\toprule
\multirow{2.5}{*}{Method} & \multicolumn{3}{c}{CIFAR100  ($B$=50)} && \multicolumn{3}{c}{ImageNet100 ($B$=50)} && \multicolumn{2}{c}{ImageNet ($B$=100)}  \\
\cmidrule{2-4}\cmidrule{6-8}\cmidrule{10-11}
& $S$=10 & 5 &  2 &&  10 &  5 &  2 &&  100 &  50 \\
\midrule
LwF\cite{li2017learning} & 53.59\scriptsize{$\pm$0.51} & 48.66\scriptsize{$\pm$0.58} & 45.56\scriptsize{$\pm$0.28} && 53.62$^{\dagger}$ & 47.64$^{\dagger}$ & 44.32$^{\dagger}$ && 40.86\scriptsize{$\pm$0.13} & 27.72\scriptsize{$\pm$0.12} \\
iCaRL\cite{rebuffi2017icarl} & 60.82\scriptsize{$\pm$0.03} & 53.74\scriptsize{$\pm$0.25} & 47.86\scriptsize{$\pm$0.41} && 65.44$^{\dagger}$ & 59.88$^{\dagger}$ & 52.97$^{\dagger}$ && 49.56\scriptsize{$\pm$0.09} & 42.61\scriptsize{$\pm$0.15} \\
BiC\cite{wu2019large} & 51.58\scriptsize{$\pm$0.16} & 48.07\scriptsize{$\pm$0.02} & 43.10\scriptsize{$\pm$0.37} && 70.07$^{\dagger}$ & 64.96$^{\dagger}$ & 57.73$^{\dagger}$ && 43.23\scriptsize{$\pm$0.13} & 38.83\scriptsize{$\pm$0.12} \\
\midrule
LUCIR\cite{hou2019learning} & 66.27\scriptsize{$\pm$0.28} & 60.80\scriptsize{$\pm$0.29} & 52.96\scriptsize{$\pm$0.25} && 70.60\scriptsize{$\pm$0.43} & 67.76\scriptsize{$\pm$0.40} & 62.76\scriptsize{$\pm$0.22} && 56.40\scriptsize{$\pm$0.10} & 52.75\scriptsize{$\pm$0.18} \\
\ \ \cellcolor{hl}{+CwD (ours)} &
\cellcolor{hl}{67.26}\scriptsize{$\pm$0.16} & \cellcolor{hl}{62.89}\scriptsize{$\pm$0.09} & \cellcolor{hl}{56.81}\scriptsize{$\pm$0.21} &\cellcolor{hl}{}& \cellcolor{hl}{71.94}\scriptsize{$\pm$0.11} & \cellcolor{hl}{69.34}\scriptsize{$\pm$0.31} & \cellcolor{hl}{65.10}\scriptsize{$\pm$0.59} &\cellcolor{hl}{}& \cellcolor{hl}{57.42}\scriptsize{$\pm$0.11} & \cellcolor{hl}{53.37}\scriptsize{$\pm$0.22} \\
\midrule
PODNet\cite{douillard2020podnet} & 66.98\scriptsize{$\pm$0.13} & 63.76\scriptsize{$\pm$0.48} & 61.00\scriptsize{$\pm$0.18} && 75.71\scriptsize{$\pm$0.37} & 72.80\scriptsize{$\pm$0.35} & 65.57\scriptsize{$\pm$0.41} && 57.01\scriptsize{$\pm$0.12} & 54.06\scriptsize{$\pm$0.09} \\
\ \ \cellcolor{hl}{+CwD (ours)} &
\cellcolor{hl}{67.44}\scriptsize{$\pm$0.35} & \cellcolor{hl}{64.64}\scriptsize{$\pm$0.38} & \cellcolor{hl}{62.24}\scriptsize{$\pm$0.32} &\cellcolor{hl}{}& \cellcolor{hl}{76.91}\scriptsize{$\pm$0.10} & \cellcolor{hl}{74.34}\scriptsize{$\pm$0.02} & \cellcolor{hl}{67.42}\scriptsize{$\pm$0.07} &\cellcolor{hl}{}& \cellcolor{hl}{58.18}\scriptsize{$\pm$0.20} & \cellcolor{hl}{56.01}\scriptsize{$\pm$0.14} \\
\midrule
AANet\cite{liu2021adaptive} & 69.79\scriptsize{$\pm$0.21} & 67.97\scriptsize{$\pm$0.26} & 64.92\scriptsize{$\pm$0.30} && 71.96\scriptsize{$\pm$0.12} & 70.05\scriptsize{$\pm$0.63} & 67.28\scriptsize{$\pm$0.34} && 51.76$^{*}$\scriptsize{$\pm$0.14} & 46.86$^{*}$\scriptsize{$\pm$0.13} \\
\ \ \cellcolor{hl}{+CwD (ours)} &
\cellcolor{hl}{70.30}\scriptsize{$\pm$0.37} & \cellcolor{hl}{68.62}\scriptsize{$\pm$0.17} & \cellcolor{hl}{66.17}\scriptsize{$\pm$0.13} &\cellcolor{hl}{}& \cellcolor{hl}{72.92}\scriptsize{$\pm$0.29} & \cellcolor{hl}{71.10}\scriptsize{$\pm$0.16} & \cellcolor{hl}{68.18}\scriptsize{$\pm$0.27} &\cellcolor{hl}{}& \cellcolor{hl}{52.30$^{*}$}\scriptsize{$\pm$0.08} & \cellcolor{hl}{47.61$^{*}$}\scriptsize{$\pm$0.20} \\
\bottomrule
\end{tabular}
\caption{\textbf{Comparison of average incremental accuracy (\%) with or without Class-wise Decorrelation (CwD) at the initial phase.}
$B$ denotes number of classes learned at initial phase and $S$ denotes number of classes learned per phase after the initial one.
Number of exemplars for each class is 20.
For AANet, we use its version based on LUCIR\cite{hou2019learning}. AANet\cite{liu2021adaptive} on ImageNet (denoted by $^{*}$) is running without class-balance finetuning after each phase due to missing implementation in its code.
All results are (re)produced by us except the ones denoted by $^{\dagger}$, which are taken from \cite{liu2021adaptive}. Results (re)produced by us are averaged over 3 runs (mean$\pm$std).}
\vspace{-3mm}
\label{all_improves}
\end{table*}
}
}
\section{Experiments}
\label{experiment}
In this section, we first elaborate on the experimental setups in Sec.~\ref{settings}.
Next, in Sec.~\ref{sota}, we add our proposed Class-wise Decorrelation (CwD) on some State-Of-the-Art (SOTA) methods \cite{hou2019learning,douillard2020podnet,liu2021adaptive} to validate its effectiveness.
Finally, in Sec.~\ref{ablation}, we provide ablation study on how factors such as number of classes of the initial phase, number of exemplars for each class, and CwD coefficient ($\eta$ in Eqn.~\eqref{overall_obj}) affects our proposed method. In addition, empirical comparison between our CwD and other decorrelation methods are given in the Appendix.

\subsection{Settings}
\label{settings}
\textbf{Datasets:} We briefly introduce the 3 datasets used in our experiments: CIFAR100\cite{krizhevsky2009learning} contains $100$ classes, with $60000$ samples, and the size of each image is $32 \times 32$; ImageNet\cite{deng2009imagenet} contains $1000$ classes, with $\approx 1.3$ million samples, and the size of each image is $224 \times 224$; ImageNet100 is a $100$-class subset of the full ImageNet\cite{deng2009imagenet}, which is generated similarly  as in \cite{hou2019learning,douillard2020podnet,liu2021adaptive}.
The classes of all datasets are first shuffled using seed 1993 as in \cite{rebuffi2017icarl,hou2019learning,liu2020mnemonics,liu2021adaptive,simon2021learning,hu2021distilling}, and then split into multiple phases.

\textbf{Implementation Details:}
For all experiments, we use ResNet18\cite{he2016deep} and the SGD optimizer, with the batch size of $128$. The strategy of Herding\cite{rebuffi2017icarl,hou2019learning,douillard2020podnet,liu2021adaptive} is used to select exemplars after each phase.
For experiments based on CIFAR100, in each CIL phase, all models are trained for $160$ epochs and the learning rate is divided by $10$ at the 80-th and  120-th epoch.
For experiments based on ImageNet100/ImageNet, in each phase, all models are trained for $90$ epochs and the learning rate is divided by $10$ at the 30-th and  60-th epoch.

When performing CIL, we reserve a constant number of exemplars for each previously learned class as had been done in  \cite{hou2019learning,douillard2020podnet,liu2020mnemonics,liu2021adaptive}.

\textbf{Baselines:}
We apply our proposed CwD to the following three strong SOTA baselines: LUCIR\cite{hou2019learning}, PODNet\cite{douillard2020podnet}, and AANet\cite{liu2021adaptive}.
For AANet, we use its version based on LUCIR\cite{hou2019learning}.
In addition, we also report results of some other classical methods, including LwF\cite{li2017learning}, iCaRL\cite{rebuffi2017icarl} and BiC\cite{wu2019large} for comparison.

\textbf{Evaluation Metric:}
We use the the average incremental accuracy to evaluate performance of CIL methods as in \cite{hou2019learning,douillard2020podnet,liu2020mnemonics,liu2021adaptive,hu2021distilling}.
Formally, suppose the CIL is conducted for $N+1$ phases and test accuracy at the $i$-th phase is $A_{i}$, then the average increment accuracy is defined as:
\begin{equation}
\label{eval_metric}
    \bar{A} = \frac{1}{N+1}\sum_{i=0}^{N}A_{i}.
\end{equation}

\begin{figure*}
    \centering
    \includegraphics[width=17cm]{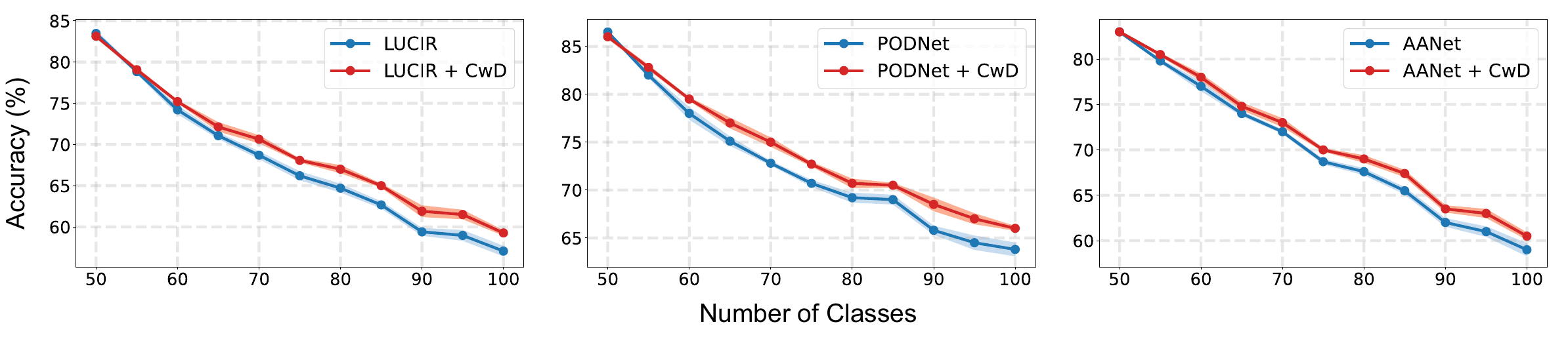}
    \vspace{-4mm}
    \caption{\textbf{Accuracy at each phase.} With ImageNet100, learning 50 classes at initial phase and 5 classes per phase for the rest 50 classes. Shading areas denote standard deviation.}
    \label{perf_plot}
    \vspace{-4mm}
\end{figure*}

\subsection{CwD improves previous SOTA methods}
\label{sota}
In this section, we add our proposed CwD on three previous SOTA methods, namely LUCIR\cite{hou2019learning}, PODNet\cite{douillard2020podnet} and AANet\cite{liu2021adaptive}, to validate the effectiveness of our method.
We denote the number of classes learned at initial phase by $B$ and number of new classes learned per phase after the initial one by $S$.
For CIFAR100 and ImageNet100, we evaluate our methods with three protocols: learning with $B=50$ at the initial phase, and then learning with $S =10/5/2$ for the rest of the classes.
For ImageNet, we evaluate our methods with two protocols: learning with $B=100$ at the initial phase, and then learning with $S=100/50$ for the rest of the classes.
The number of exemplars of each class is $20$ for these experiments.
The experimental results are shown in Tab.~\ref{all_improves}.

As shown in Tab.~\ref{all_improves}, across various CIL benchmarks, CwD can consistently improve the average incremental accuracy of previous SOTAs by around 1\% to 3\%.
In particular, as we increment with fewer classes per phase (\ie, smaller $S$), the improvements provided by adding CwD at the initial phase is more apparent.
For example, with LUCIR\cite{hou2019learning} on CIFAR100 and ImageNet100 when $S =2$, CwD can improve the baseline performance by up to 3.85\% and 2.34\%, respectively.
This shows that, in settings that are more challenging (\eg, with longer incremental learning sequences), our CwD can be even more effective.

Furthermore, in the large-scale CIL setting of ImageNet, CwD enables the weaker baseline LUCIR\cite{hou2019learning} to compete with or even outperform the stronger baseline  PODNet~\cite{douillard2020podnet}.

To further understand CwD's improvement at each phase, we plot how each method's accuracy at each phase (\ie, $A_{i}$ in Eqn.~\eqref{eval_metric}) changes as CIL proceeds on ImageNet100 with $S =5$.
From Fig.~\ref{perf_plot}, we observe similar situation as directly mimic oracle model representation (see Fig.~\ref{oracle_results}): our CwD barely affects accuracy at the initial phase, but bring strong improvements when learning at the following phases.
This shows that the improvements by CwD on average incremental accuracy is not due to na\"ive performance improvements on the initial phase, but mainly due to an initial-phase representation that is favorable for incrementally learning new classes.

\setlength{\tabcolsep}{2mm}{
\renewcommand{\arraystretch}{1.15}{
\begin{table}[ht]
\centering
\begin{tabular}{l|cccc}
\toprule
$S$ & $B$ & LUCIR & +CwD (ours) &$\uparrow$ \\
\midrule
\multirow{5}{*}{$10$} & 10 & 57.01\scriptsize{$\pm$0.14} & 57.90\scriptsize{$\pm$0.07} & +0.89	\\
& 20 & 61.21\scriptsize{$\pm$0.35} & 62.49\scriptsize{$\pm$0.36} & +1.28\\
& 30 & 64.82\scriptsize{$\pm$0.38} & 66.54\scriptsize{$\pm$0.35} & +1.72\\
& 40 & 67.68\scriptsize{$\pm$0.37} & 69.70\scriptsize{$\pm$0.10} & +2.02\\
& 50 & 70.60\scriptsize{$\pm$0.43} & 71.94\scriptsize{$\pm$0.11} & +1.33\\
\midrule
\multirow{5}{*}{$5$} & 10 & 50.47\scriptsize{$\pm$0.31} & 51.92\scriptsize{$\pm$0.10} & +1.45 \\
& 20 & 56.41\scriptsize{$\pm$0.37} & 58.14\scriptsize{$\pm$0.13} & +1.73 \\
& 30 & 61.00\scriptsize{$\pm$0.09} & 63.18\scriptsize{$\pm$0.14} & +2.18 \\
& 40 & 63.73\scriptsize{$\pm$0.23} & 66.25\scriptsize{$\pm$0.16} & +2.52 \\
& 50 & 67.76\scriptsize{$\pm$0.40} & 69.34\scriptsize{$\pm$0.31} & +1.58 \\

\bottomrule
\end{tabular}
\caption{\textbf{Ablation study on impact of number of classes of the initial phase.}
In these experiments, we learn $B$ classes at the initial phase, and then learn the rest of the classes with $S$ classes per phase. Number of exemplars is set to 20 for each class. Average incremental accuracy (\%) is reported.
All results (mean$\pm$std) are produced by us, and are averaged over 3 runs.}
\label{ablation_init}
\vspace{-3mm}

\end{table}
}
}

\subsection{Ablation study}
\label{ablation}
In this section, we study how the following factors affect the effectiveness of CwD: (1) number of classes for the initial phase, (2) number of exemplars, and (3) CwD coefficient ($\eta$ in Eqn.~\eqref{overall_obj}). Experiments in this section are based on LUCIR\cite{hou2019learning}, with the ResNet18 model and the ImageNet100 dataset.

\textbf{Ablation on the number of classes of the initial phase.} We vary the number of classes learned at the initial phase (denoted by $B$) from 10 to 50 and run LUCIR\cite{hou2019learning} with or without our CwD. 
From the results in Tab.~\ref{ablation_init}, we can see that, even when $B =10$, CwD can bring a notable improvements (0.89\% when $S =10$ and 1.45\% when $S =5$).
This shows that our CwD can improve CIL even when there is only a few number of classes in the initial phase. Note that this setting is quite challenging, since CwD can only directly affect representations of 10 out of 100 classes.

Additionally, as $B$ increases, the performance of CwD first increases and reaches the peak when $B =40$ (2.02\% and 2.52\% when $S =10$ and $5$ respectively), and then decreases.
We posit that the reason for this increasing-then-decreasing trend is that, as $B$ increases from $10$ to~$40$, CwD can directly affect representations of more classes (in the initial phase), which enables it to yield greater improvements.
However, as $B$ continues increasing from $40$ to $50$, the gap between CIL and jointly training with all classes shrinks, resulting in less room for improvement of CIL.

\setlength{\tabcolsep}{2mm}{
\renewcommand{\arraystretch}{1.15}{
\begin{table}[ht]
\centering
\begin{tabular}{l|cccc}
\toprule
$S$ & $R$ & LUCIR & +CwD (ours) &$\uparrow$ \\
\midrule
\multirow{5}{*}{$10$} & 40 & 72.41\scriptsize{$\pm$0.61} & 73.29\scriptsize{$\pm$0.11} & +0.88 \\
& 30 & 71.70\scriptsize{$\pm$0.37} & 72.63\scriptsize{$\pm$0.15} & +0.93 \\
& 20 & 70.60\scriptsize{$\pm$0.43} & 71.94\scriptsize{$\pm$0.11} & +1.34 \\
& 10 & 68.73\scriptsize{$\pm$0.52} & 69.77\scriptsize{$\pm$0.04} & +1.04 \\
& 5 & 66.49\scriptsize{$\pm$0.52} & 67.63\scriptsize{$\pm$0.07} & +1.14\\

\midrule
\multirow{5}{*}{$5$} & 40 & 70.74\scriptsize{$\pm$0.49} & 72.06\scriptsize{$\pm$0.11} & +1.32 \\
& 30 & 68.56\scriptsize{$\pm$0.42} & 70.04\scriptsize{$\pm$0.12} & +1.44 \\
& 20 & 67.76\scriptsize{$\pm$0.40} & 69.34\scriptsize{$\pm$0.31} & +1.58 \\
& 10 & 64.07\scriptsize{$\pm$0.38} & 66.07\scriptsize{$\pm$0.46} & +2.00 \\
& 5 & 60.41\scriptsize{$\pm$0.77} & 62.58\scriptsize{$\pm$0.53} & +2.17 \\
\bottomrule
\end{tabular}
\vspace{-2mm}
\caption{\textbf{Ablation study on impact of number of exemplars.}
In these experiments, we learn 50 classes at the initial phase, and then learn the rest of the classes with $S$ classes per phase. Number of exemplars for each class is $R$. Average incremental accuracy (\%) is reported. All results (mean$\pm$std) are produced by us, and are averaged over 3 runs.}
\label{ablation_replay}
\vspace{-4mm}
\end{table}
}
}

\textbf{Ablation on the number of exemplars.}
In this ablation, we vary the number of exemplars per class (denoted by $R$) and run LUCIR\cite{hou2019learning} with or without CwD. 

As can be observed in Tab.~\ref{ablation_replay}, with more exemplars per class (\eg, $R =40$), the gap between CIL and jointly training with all classes is smaller, resulting in a still notable but smaller improvements (0.88\% when $S =10$ and 1.32\% when $S=5$).
As $R$ decrease, improvements provided by our CwD generally increase, reaching 1.34\% when $S =10$ and $R =20$, and 2.17\% when $S =5$ and $R =5$.
This shows that our CwD yields more significant improvements in the more challenging setting with less exemplars.

\begin{figure}
    \centering
    \includegraphics[width=7.5cm]{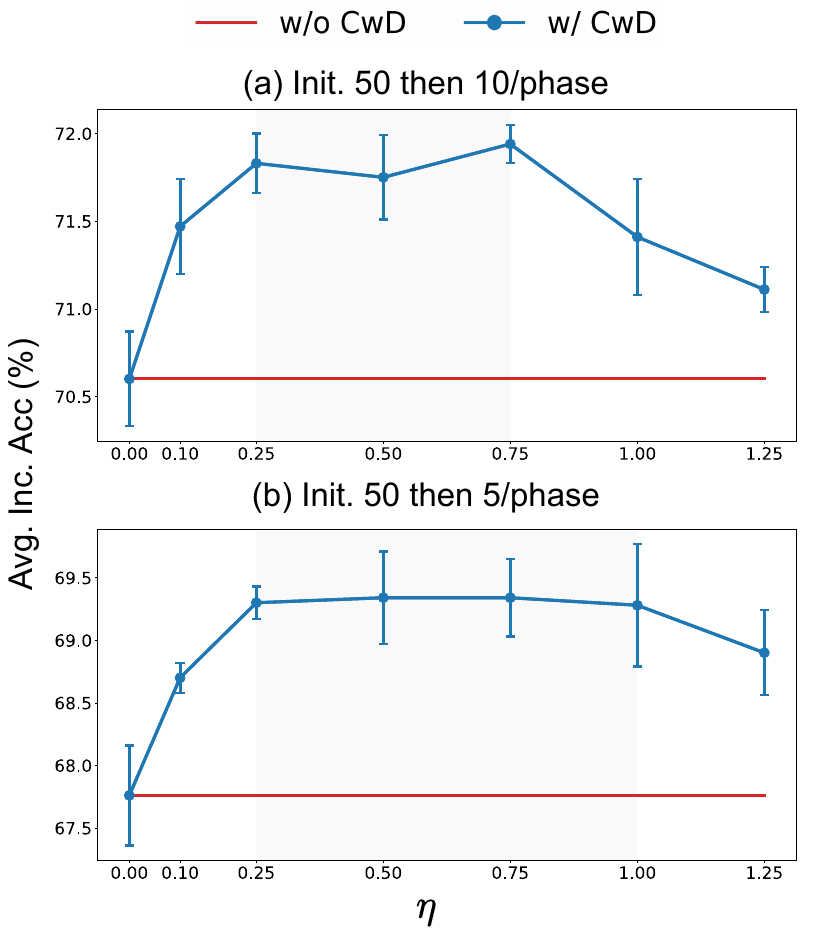}
    \vspace{-4mm}
    \caption{\textbf{Impact of CwD Coefficient ($\eta$ in Eqn.~\eqref{overall_obj}).} 
    Average incremental accuracy (\%) is reported. Results are averaged over 3 runs. Shadowed region denotes intervals that CwD is robust to $\eta$.
    }
    \label{aux_ablate}
    \vspace{-5mm}
\end{figure}

\textbf{Ablation on the CwD coefficient.}
In this ablation, we study how the average incremental accuracy changes when CwD coefficient (\ie, $\eta$ in Eqn.~\eqref{overall_obj}) varies in the range of $[0, 1.25]$. This ablation is based on LUCIR\cite{hou2019learning} and ImageNet100, with two CIL protocols that initially learn 50 classes and then learn 10/5 classes per phase for the rest.

From Fig.~\ref{aux_ablate}, we can see that, with both protocols, as $\eta$ increases, the performance improvement brought by CwD first increases, then plateaus, and finally decrease.

The stable performance improvement (\ie, the plateau interval) suggests that our CwD is robust to the choice of $\eta$.
Furthermore, the plateau interval of the second protocol is larger than that of the first protocol ($[0.25, 1.0]$ vs $[0.25, 0.75]$), indicating that our CwD is even more robust to $\eta$ when the incremental learning sequence is longer.

When the $\eta$ is too large (\eg, larger than $0.75$ for the first protocol and $1.0$ for the second), the performance gain demonstrates a decreasing trend.
The reason might be that an excessively large penalization on the Frobenius norm of class-wise correlation matrix would make the representations of each class to spread out drastically, resulting in large overlaps among representations of different classes. This is further studied in the Appendix.

\section{Conclusion}
In this work, we study Class Incremental Learning (CIL) from a previously underexplored viewpoint --- improving CIL by mimicking the oracle model representation at the initial phase. Through extensive experimental analysis, we show the tremendous potential of this viewpoint.  We propose a novel CwD regularization term for improving the representation of the initial phase. Our CwD regularizer yields consistent and significant performance improvements over three previous SOTA methods across multiple benchmark datasets with different scales.

Although we have developed our method through observations on the differences between the oracle model and the na\"ively-trained initial-phase model, the underlying reason why more uniformly scattered representations for each class benefit CIL is still yet to be explored. 
Some analysis on why CwD improves CIL are provided in the Appendix, but we leave more rigorous understandings as future works.

\section*{Acknowledgement}
The authors thank anonymous reviewers for their constructive feedback.
Yujun Shi and Vincent Tan are funded by a Singapore National Research Foundation (NRF) Fellowship
(R-263-000-D02-281) and a Singapore Ministry of Education AcRF Tier 1 grant (R-263-000-E80-114). Jian Liang is funded by Beijing Nova Program (Z211100002121108). Philip Torr is funded by Turing AI Fellowship EP/W002981/1.

{\small
\bibliographystyle{ieee_fullname}
\bibliography{egbib}
}
\clearpage
\onecolumn
\appendix
\section{Proof of Proposition 1 (Main text Sec.~2.3)}


\begin{proof}
Recall that for a $d$-by-$d$ correlation matrix $K$, we have:
\begin{equation}
\label{eigen_trace}
    \sum_{i=1}^{d}\lambda^{(c)}_{i} = \tr(K) = d.
\end{equation}
This is because for a square matrix, summation of eigenvalues equals trace of the matrix. In addition, for a correlation matrix, all its diagonal elements are 1, which results in $\
tr(K) = d$.

Next, for the left hand side of the equation, we have:
\begin{equation}
\begin{aligned}
\sum_{i=1}^{d}(\lambda_{i} - \frac{1}{d}\sum_{j=1}^{d}\lambda_{j})^{2} &= \sum_{i=1}^{d}(\lambda_{i} - 1)^{2} &\text{(Plug-in Eqn.~\eqref{eigen_trace})}\\
&= \sum_{i=1}^{d}\lambda^{2}_{i} - 2\sum_{i=1}^{d}\lambda_{i} + d \\
&= \sum_{i=1}^{d}\lambda^{2}_{i} - d &\text{(Plug-in Eqn.~\eqref{eigen_trace})}.
\end{aligned}
\end{equation}
Next, for the right hand side, we have:
\begin{equation}
\begin{aligned}
   \|K\|_{\mathrm{F}}^{2} -d &= \tr(K^{T}K) - d \\ 
   &= \tr(U\Sigma U^{T}U\Sigma U^{T}) - d &\text{(Applying eigendecomposition on $K$)} \\
   &= \tr(U\Sigma^{2}U^{T}) - d \\
   &= \sum_{i=1}^{n}\lambda_{i}^{2} - d.
\end{aligned}
\end{equation}
Therefore, we have shown that left hand side of the equation equals the right hand side.
\end{proof}

\section{More Analysis on Why CwD Improves CIL}
\label{why_cwd_works}
Although we have developed our method through observations on the differences between the oracle model and the na\"ively-trained initial-phase model, the underlying reason why more uniformly scattered representations for each class benefit CIL is still yet to be explored.

In this section, we provide additional analysis on the effectiveness of CwD. We posit the reason why CwD improves CIL is as follow: after applying CwD, data representations produced by the model are not overly compressed. That being said, the representations will contain more information about input data besides the information useful for classification at the initial phase.
Although these additional information are not useful for classification at the initial phase, they can be useful for discriminating between classes of the initial phase and future phases, which will improve CIL when incrementally learning new classes.

\begin{figure}
    \centering
    \includegraphics[width=12cm]{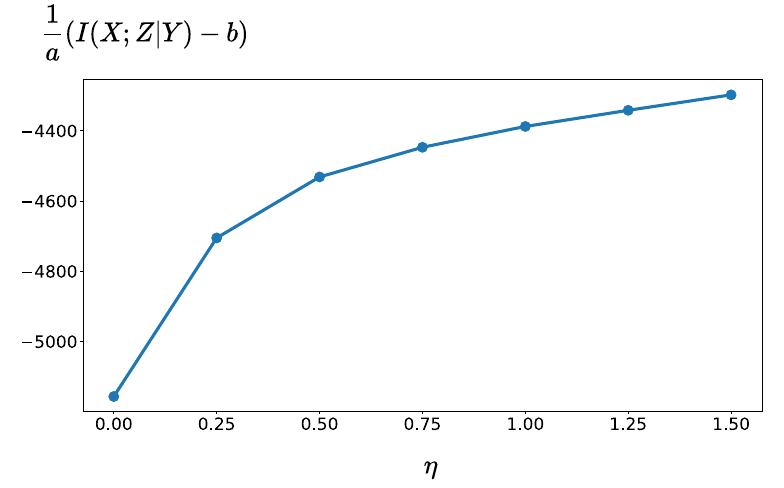}
    \caption{\textbf{Visualization on how $I(X;Z|Y)$ change with increasing $\eta$.} X-Axis is $\eta$. Y-Axis is $\frac{1}{C}\sum_{c=1}^{C}\sum_{i=1}^{d}\log\lambda^{(c)}_{i} = \frac{1}{a}(I(X;Z|Y)-b)$ as derived in Eqn.~\eqref{cond_mi_est}. Note that here Y-Axis is negative because it is conditional mutual information minus a very large positive constant.}
    \label{info_inc}
\end{figure}

To quantitatively validate that CwD helps preserve more information in representations beyond classification at initial phase, we adopt the information theoretic quantity $I(X;Z|Y)$, where random variable $X$ is input data, $Z$ is the representation produced by model given $X$, and $Y \in \{1,2,...,C\}$ is the label of $X$. $I(X;Z|Y)$ is the conditional mutual information, which characterizes, how much information is left in $Z$ about $X$ when   conditioned on knowing $Y$. 
Therefore, larger $I(X;Z|Y)$ implies that more information about $X$ are preserved in $Z$ besides the purpose of discriminating classes in $\{1,2,...,C\}$.

In addition, following the exposition in the main text, we denote the covariance matrix of $Z|Y=c$ as $K^{(c)}$. The number of classes at the initial phase is denoted by $C$. We view $Z$ and $X$ as discrete random variable\footnote{This is reasonable because as we know, each pixel of an image is represented by $3\times8=24$ bits, which makes an image a discrete random variable.}.
In addition, we assume that \textbf{$P_{Z|Y}(z|c)$ is a quantized Gaussian distribution}. That is to say, although $Z|y=c$ is a discrete random variable, its probability mass function is infinitely close to the density function of a \textbf{Gaussian distribution $p_{Z|Y}(z|c)$}. Therefore, we shall have: $P_{Z|Y}(z|c) = p_{Z|Y}(z|c)\delta$.

To start with, we study $I(X;Z|Y=c)$, which is the mutual information between $X$ and $Z$ conditioned on $Y$ being a specific class $c$:
\begin{equation}
\label{cond_mi_est_c}
\begin{aligned}
    I(X;Z|Y=c) &= H(Z|Y=c) - H(Z|X, Y=c) \\
    &= H(Z|Y=c) + 0 \;\;\;\;\;\text{($Z$ is deterministic conditioned on $X$.)} \\
    &= \sum_{z\in\mathcal{Z}} P_{Z|Y}(z|c)\delta\log{\frac{1}{P_{Z|Y}(z|c)\delta}} \\
    &\approx \int_{\mathcal{Z}} -p_{Z|Y}(z|c) \log p_{Z|Y}(z|c) + \log{\frac{1}{\delta}} \\
    &= a\log\det(K^{(c)}) + b \;\;\;\;\;\text{(closed-form solution of differential entropy on Gaussian variables.)}\\
    &= a\sum_{i=1}^{d}\log\lambda^{(c)}_{i} + b,
\end{aligned}
\end{equation}
where $a$, $b$ are positive constant.
Furthermore, we assume that probability of one sample being any class are the same, which further leads to:
\begin{equation}
\label{cond_mi_est}
\begin{aligned}
    I(X;Z|Y) &= \sum_{c=1}^{C} p(Y=c)I(X;Z|Y=c) \\
             &=a\frac{1}{C}\sum_{c=1}^{C}\sum_{i=1}^{d}\log\lambda^{(c)}_{i} + b.
\end{aligned}
\end{equation}
To this end, we have obtained an estimation on $I(X;Z|Y)$ given Eqn.~\eqref{cond_mi_est}.

Based on what we have derived in Eqn.~\eqref{cond_mi_est}, we can visualize $\frac{1}{C}\sum_{c=1}^{C}\sum_{i=1}^{d}\log\lambda^{(c)}_{i}$ as an proxy of $I(X;Z|Y)$. In this way, we can get to know how $I(X;Z|Y)$ varies for model trained with different CwD coefficient (i.e., $\eta$ in eq.~(9) of main text). Results are shown in Fig.~\ref{info_inc}.

As can be seen, with larger $\eta$, $I(X;Z|Y)$ consistently increase. This means that as we increase $\eta$, representation $Z$ will contain more information about $X$ besides information useful for classification at initial phase. Although these additional information are not useful for classification at initial phase, they could be useful for discriminating between classes of initial phase and future phases. In this way, CwD might be beneficial for incrementally learning new classes.

We leave more systematic and rigorous studies on why CwD can help CIL as future works.

\section{Analysis on Why CwD Coefficient Being Too Large is Bad (Main text Sec.~3.3)}
In the third ablation study of Sec.~3.3 of main text, we ablate how the CwD Coefficient (i.e., $\eta$ in eq.~(9) of main text) affect the improvements brought by CwD. From the results, we find that as we increase $\eta$, performance gain brought by CwD will eventually decrease when $\eta$ is too large.

To understand this phenomenon, we define the volume occupied by representations of class $c$ as $V^{(c)}$. By assuming representations of class $c$ are Gaussian distributed, we have:
\begin{equation}
\label{volumn}
    V^{(c)} \propto \prod_{i=1}^{d} \lambda^{(c)}_{i},
\end{equation}
where $(\lambda^{(c)}_{1},...,\lambda^{(c)}_{d})$ are eigenvalues of the covariance matrix of representations of class $c$. This is based on the geometric interpretation of eigenvalues of covariance matrix.
Further, combining Eqn.~\eqref{volumn} with Eqn.~\eqref{cond_mi_est_c}, we have:
\begin{equation}
\label{connection_v_mi}
    \log V^{(c)} \propto \sum_{i=1}^{d}\log\lambda_{i}^{(c)} \propto I(X;Z|Y=c).
\end{equation}
Based on Eqn.~\eqref{connection_v_mi} and the observation in Sec.~\ref{why_cwd_works} that $I(X;Z|Y=c)$ consistently increase with increasing $\eta$, we know that representations of class $c$ will occupy more volume in representation space when $\eta$ increase. Therefore, when $\eta$ is too large, representations of initial phase classes will occupy too much space, causing large overlaps with classes in future phases.

\begin{figure}[htbp]
    \centering
    \includegraphics[width=12cm]{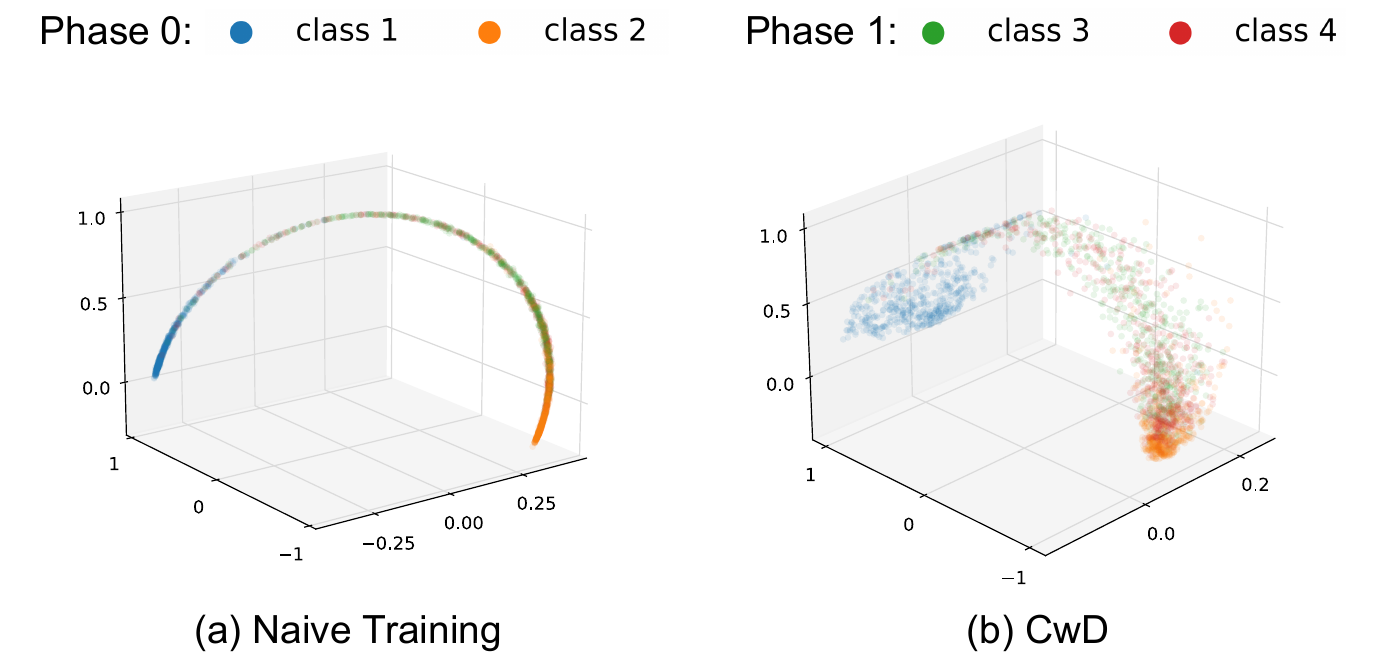}
    \caption{Visualization of representations (normalized to the unit sphere) in a two-phase CIL setting (learning 2 classes for each phase).
    \textbf{(a)} Na\"ive training at the initial phase (a.k.a., phase~$0$) but visualize on data of both phase~$0$ and phase~$1$ (i.e., including all 4 classes). The model has not been trained on class 3 and class 4, but data representations of these class still lie in a long and narrow region between representations of class 1 and class 2.
    \textbf{(b)} Training with CwD at the initial phase (a.k.a., phase~$0$) but visualize on data of both phase~$0$ and phase~$1$. With our CwD, not only representations of class 1 and class 2 scatter more uniformly, representations of class 3 and class 4 also scatter more uniformly in the space.
    }
    \label{fig1_supp}
\end{figure}

\section{Extended Visualization on Figure 1 of Main Text}
\textbf{More Details on Setups in Figure 1 of Main Text.} For the two-phase CIL setting, each phase contains 2 classes from CIFAR100. We train an AlexNet model for visualization. The representation dimension of AlexNet is set to 3 for convenience of visualization. Representations are normalized to unit sphere as done in LUCIR\cite{hou2019learning}.

\textbf{Extended Visualizations.} Here, we extended visualization of Figure~1~(a) and Figure~1~(d) of main text. Although for these two figures, \textbf{the model is only trained on the 2 classes of initial phase}, we visualize data representations of all four classes. Results are shown in Fig.~\ref{fig1_supp}.

Surprisingly, from the figure, we can see that when na\"ively traiend on the 2 classes of initial phase, data representations of the other 2 classes (i.e., class 3 and class 4) also only lies in the long and narrow region between representations of class 1 and class 2. However, after applying our CwD, data representations of class 3 and class 4 are also more uniformly scattered in representation space. These phenomenon further suggest that applying our CwD can enforce representation to preserve more information about input data, and thus benefit CIL.

\begin{figure*}[htbp]
    \centering
    \includegraphics[width=17cm]{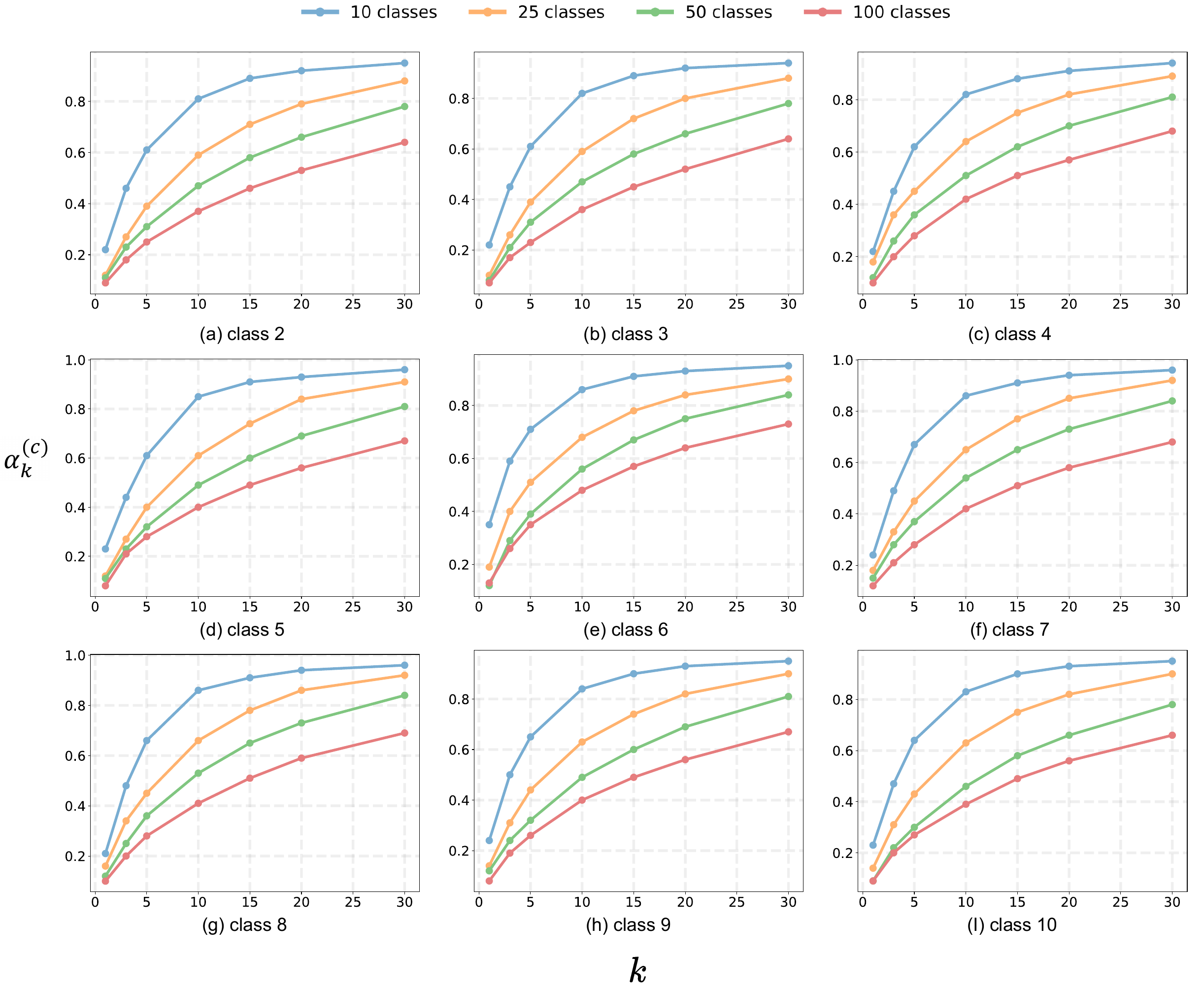}
    \caption{\textbf{Visualization on how $\alpha^{(c)}_{k}$ changes with increasing $k$ for models trained with different number of classes.} $\alpha^{(c)}_{k}$ curve generated by the model trained on 10/25/50/100 classes are denoted by blue/orange/green/red, respectively. From left to right, top to bottom are figures based on representations of class 2 to 10, respectively. The figure generated by representations of class 1 is shown in the main text.}
    \label{eigen_value_supp}
\end{figure*}
\clearpage

\section{More Visualizations on Other Classes (Main text Sec.~2.2)}
In the Sec.~2.2 of main text, based on ImageNet100, we generate four subsets containing 10/25/50/100 classes, where the subset with more classes contains the subset with fewer classes (the 10 classes of the first subset are shared by all 4 subsets). Based on these four subsets, we train four ResNet18 models. Next, given the four ResNet18 models trained on different number of classes, we visualize
$\alpha^{(c)}_{k}$ for representations of one of the 10 shared classes.
Here, we provide the visualization for the other 9 classes. The results are shown in Fig.~\ref{eigen_value_supp}. As can be observed, for the other 9 shared classes, curve of $\alpha^{(c)}_{k}$ show the similar trend as class 1 shown in the main text, i.e., for every fixed $k$, $\alpha^{(c)}_{k}$ consistently decrease as the model being trained with more classes. These results further validated our observations mentioned in main text Sec.~2.2: as the model being jointly trained with more number of classes, representations of each class scatter more uniformly.

\section{Comparison with Other Decorrelation Methods.}
As we mentioned in the main text, there are some works in other fields that utilized feature decorrelation for other purposes (e.g. boosting generalization). Here, we provide comparison between CwD and some of other decorrelation methods \cite{cogswell2015reducing,xiong2016regularizing} under CIL setting.

Specifically, based on LUCIR, we first use a ResNet18 to learn 50 classes of ImageNet100 at the initial phase, and then learn 10 new classes per phase. We apply DC\cite{cogswell2015reducing}, SDC\cite{xiong2016regularizing}, and CwD at the initial phase, respectively (see Tab.~\ref{ablation_decorr}).
Our results show that although adding DC and SDC at initial phase can also boost CIL performance due to their decorrelation effects, the improvement brought by CwD is larger. The advantage of CwD is possibly because treating each class separately can better mimic representations of oracle model.

\setlength{\tabcolsep}{0.7mm}{
\renewcommand{\arraystretch}{0.5}{
\begin{table}[ht]
\centering
\begin{tabular}{l|cccc}
\toprule
Methods & LUCIR & +DC & +SDC & \textbf{+CwD (Ours)} \\
\midrule
Acc. & 70.60\scriptsize{$\pm$0.43} & 71.39\scriptsize{$\pm$0.21} & 71.52\scriptsize{$\pm$0.40} & \textbf{71.94}\scriptsize{$\pm$0.11} \\
\bottomrule
\end{tabular}
\caption{\textbf{Comparing CwD with other decorrelation methods.} Average Incremental Accuracy (\%) is reported. All results (mean$\pm$std) are averaged over 3 runs.}
\label{ablation_decorr}
\end{table}
}
}

\section{Why not apply CwD in ``later phases''?}
\begin{figure}[htbp]
    \centering
    \includegraphics[width=10cm]{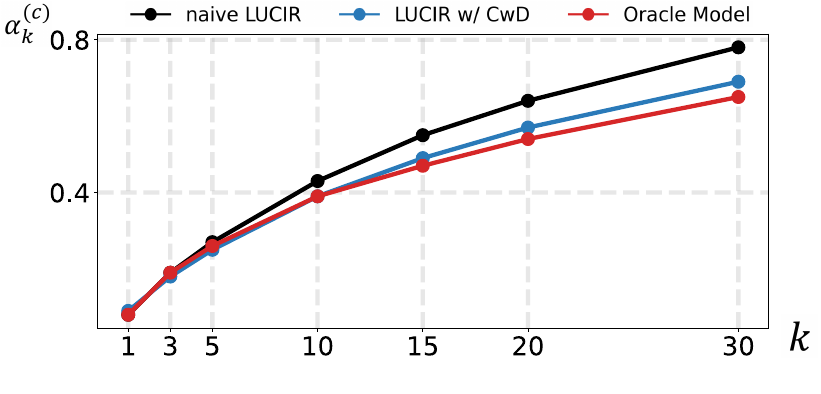}
    \caption{Visualization of  how $\alpha_{k}^{(c)}$ defined in Eqn.~(3) (of the main paper) varies with increasing $k$ for different models.}
    \label{fig_later_phase}
\end{figure}

Based on  our observations, adding CwD to the later phases is unnecessary. This is because once we apply CwD at the initial phase, representations of later phase classes will automatically be decorrelated. To see this, we apply eigenvalue analysis as in Sec.~2.2 of our paper. We analyze representations of \textbf{a class learned at a later phase} with the following three models: 1) na\"ive LUCIR model incrementally learned all classes; 2) LUCIR with CwD at initial phase; 3) oracle model. Results are shown in Fig.~\ref{fig_later_phase}. 
Our results show that even though we \textbf{only apply CwD at initial phase}, representations of the \textbf{class learned at later phases} will also mimic oracle model.

\section{What If Keep Increasing $\beta$ in Fig.~2 in Main Paper.}
\begin{figure}[H]
    \centering
    \includegraphics[width=6cm]{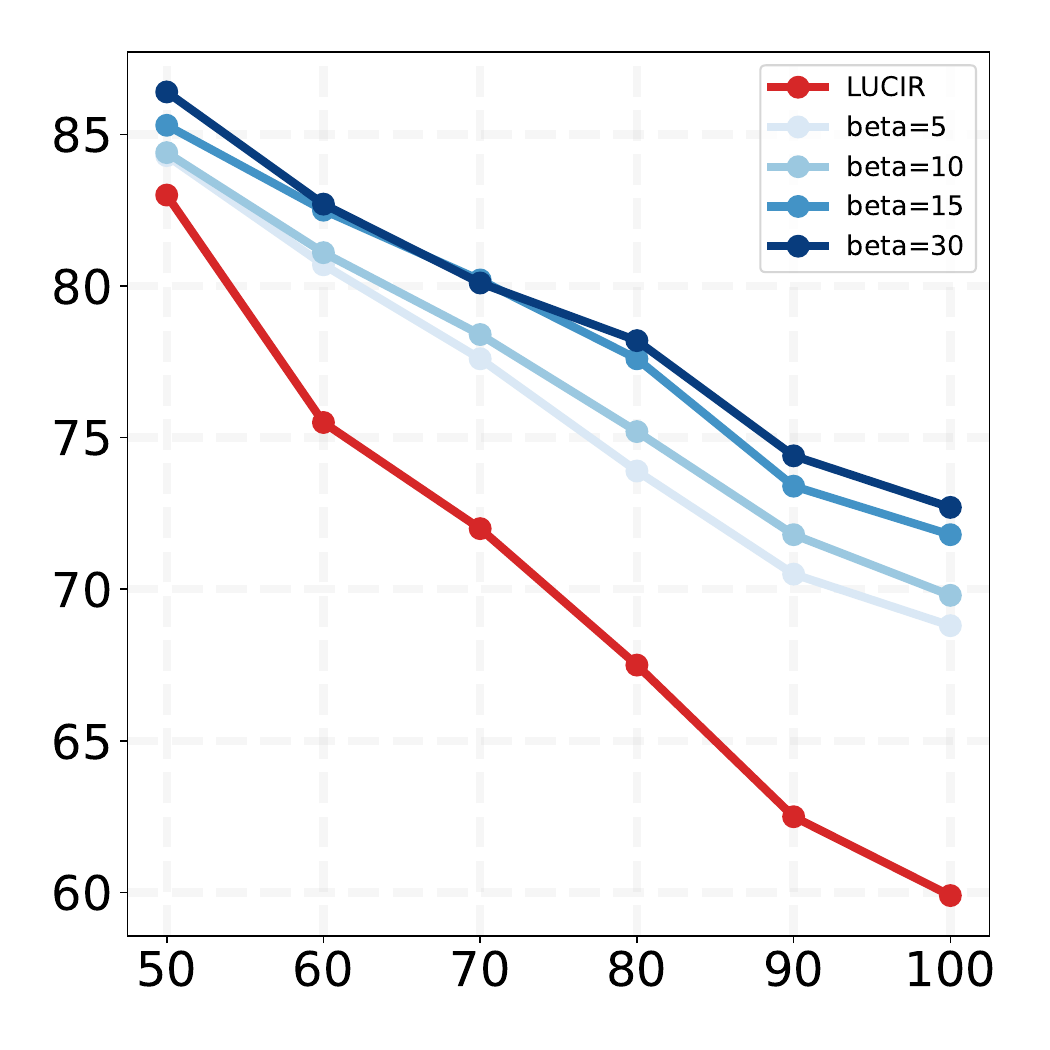}
    \includegraphics[width=6cm]{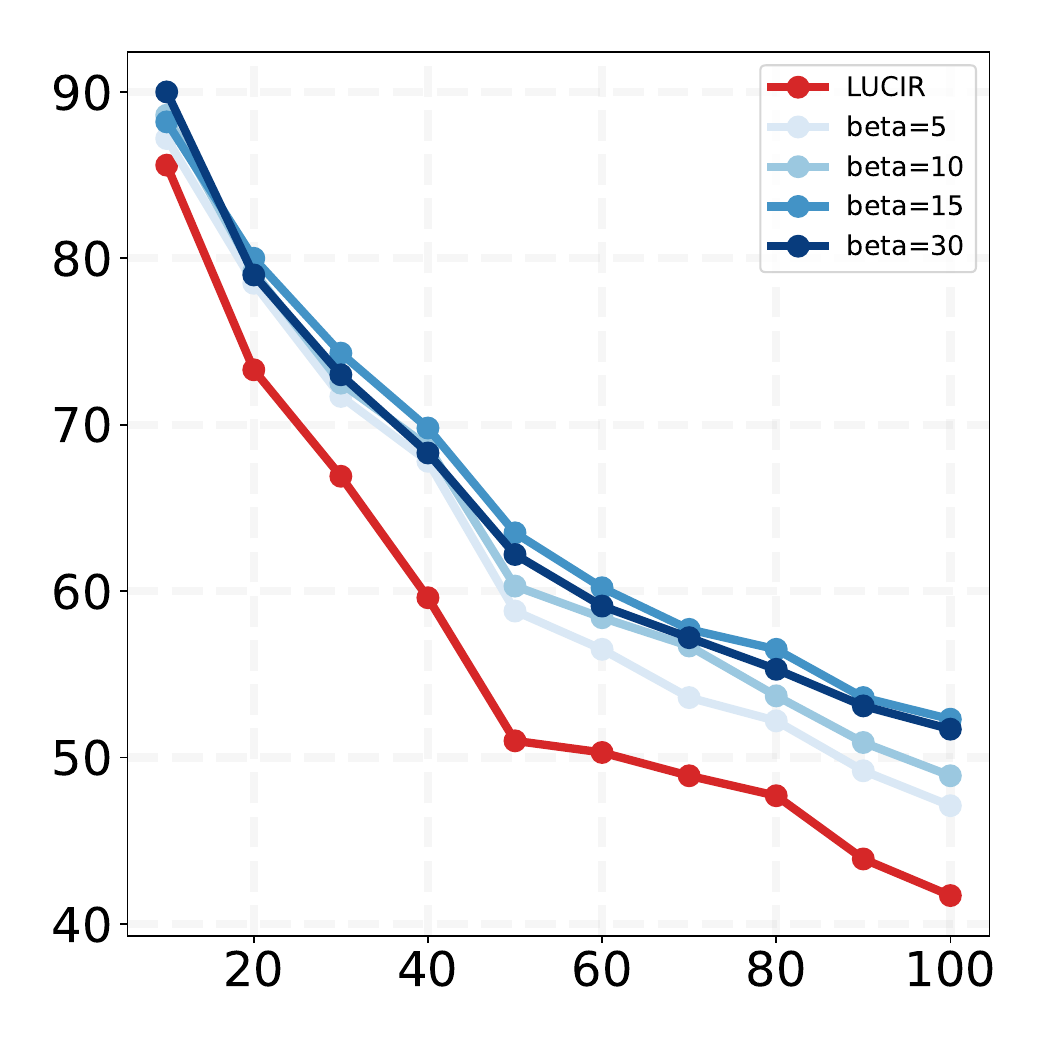}
    \caption{Exploratory experiments the same as Fig.~2 in main paper. $\beta=30$ is added comparing to Fig.~2 of main paper. As one can observe, $\beta=30$ yields approximately the same results as $\beta=15$.}
    \label{exp_appendix}
\end{figure}

According to our experiments, $\beta >15$ (e.g. $\beta=30$) yields similar performance as $\beta=15$. Results are shown in Fig.~\ref{exp_appendix}.

\end{document}